\documentclass{article}

\usepackage[preprint]{corl_2024} 

\usepackage{graphicx}
\usepackage{soul}
\usepackage{xcolor}
\usepackage{amsmath}
\usepackage{amssymb}
\usepackage{booktabs}
\usepackage{enumitem}
\usepackage{multirow}
\usepackage{array}
\usepackage{makecell}
\usepackage{floatflt}
\usepackage{wrapfig}

\newcommand{\acronym}{GraspSplats}

\title{\acronym{}: \\Efficient Manipulation with 3D Feature Splatting}

\author{
  Mazeyu Ji$^{*}$, Ri-Zhao Qiu$^{*}$, Xueyan Zou, Xiaolong Wang\\
  UC San Diego\\
  $^{*}$Equal contribution\\
  \url{https://graspsplats.github.io}
}

\begin{document}
\maketitle

\begin{abstract}
The ability for robots to perform efficient and \textbf{zero-shot} grasping of object parts is crucial for practical applications and is becoming prevalent with recent advances in Vision-Language Models (VLMs). To bridge the 2D-to-3D gap for representations to support such a capability, existing methods rely on neural fields (NeRFs) via differentiable rendering or point-based projection methods. However, we demonstrate that NeRFs are inappropriate for scene changes due to their implicitness and point-based methods are inaccurate for part localization without rendering-based optimization. To amend these issues, we propose \acronym{}. Using depth supervision and a novel reference feature computation method, \acronym{} generates high-quality scene representations in under 60 seconds. We further validate the advantages of Gaussian-based representation by showing that the explicit and optimized geometry in \acronym{} is sufficient to natively support (1) real-time grasp sampling and (2) \textbf{dynamic and articulated object manipulation} with point trackers.
With extensive experiments on a Franka robot, we demonstrate that \acronym{} significantly outperforms existing methods under diverse task settings. In particular, \acronym{} outperforms NeRF-based methods like F3RM and LERF-TOGO, and 2D detection methods.
\end{abstract}

\keywords{Zero-shot manipulation, Gaussian Splatting, Keypoint Tracking} 

\section{Introduction}

Efficient zero-shot manipulation with part-level understanding is crucial for downstream robotics applications. Consider a kitchen robot deployed to a new home: given a recipe with language instructions, the robot pulls a drawer by its handle, grasps a tool by its grips, and then pushes the drawer back. To perform these tasks, the robot must \textbf{dynamically} understand \textbf{part-level} grasp affordances to interact effectively with objects. Recent work, such as \cite{shen2023-F3RM, rashid2023-LERFTOGO, kobayashi2022-DFF}, explores this understanding by embedding reference features from large-scale pre-trained vision models (e.g., CLIP~\cite{radford2021-clip}) into Neural Radiance Fields (NeRFs). However, those methods~\cite{rashid2023-LERFTOGO,shen2023-F3RM} offer only a static understanding of the scene at the object level and require minutes to train the scene, necessitating costly retraining after any scene changes. This limitation significantly hinders practical applications involving object displacements, or requiring part-level understanding. On the other hand, point-based methods such as \cite{huang2023-voxposer}, which perform back-projection of 2D features, are efficient in feature construction but struggle with visual occlusion and often fail to infer fine-grained spatial relationships without further optimization.

\begin{figure*}[t!]
    \vspace{-10mm}
    \centering
    \includegraphics[width=\linewidth]{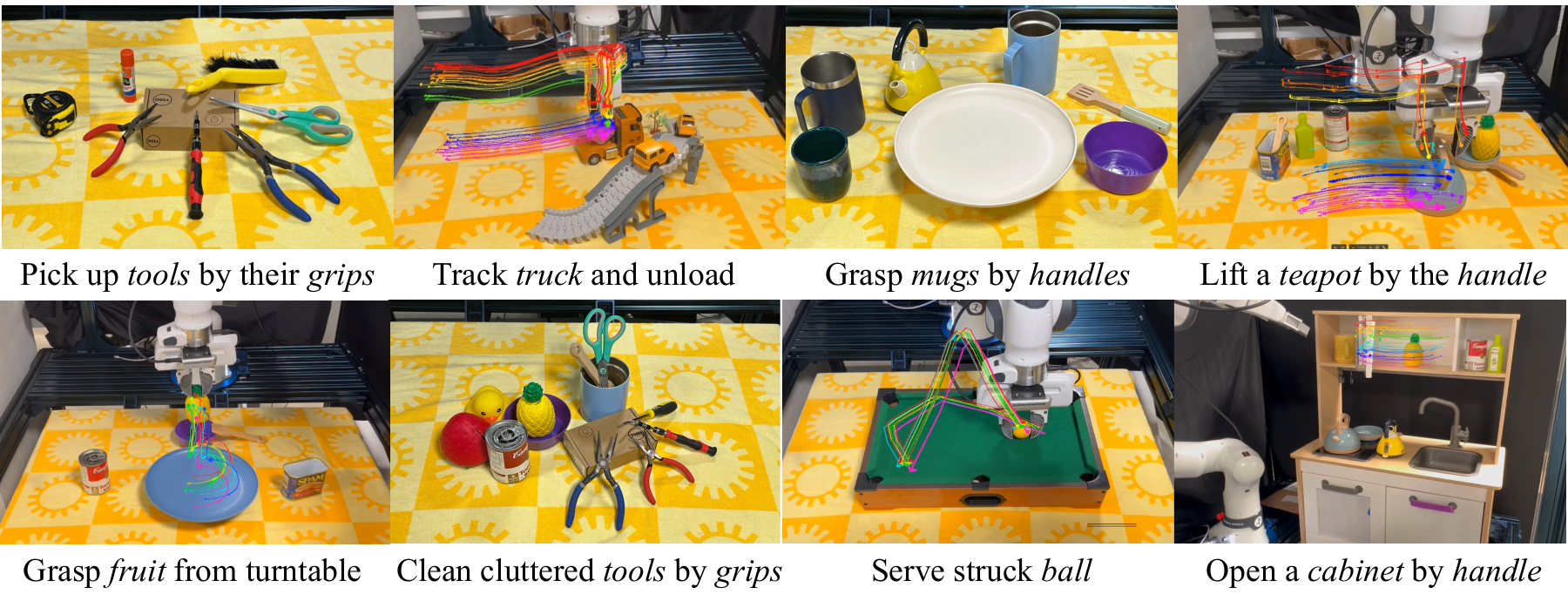}
    \caption{\textbf{\acronym{}} supports diverse robotics tasks using feature-enhanced 3D Gaussians. Compared to existing NeRF-based methods~\cite{shen2023-F3RM,rashid2023-LERFTOGO}, \acronym{} transforms the feature representation to reflect object motions in real-time with point tracking from one or more cameras, which makes it possible to perform \textbf{zero-shot dynamic and articulated object manipulation by parts}.}
    \label{fig:teaser_fig}
\end{figure*}

In addition to dynamic and part-level scene understanding, achieving fine manipulation requires the robot to have a strong understanding of both the geometry and semantics of the scene. For this capability to emerge from coarse 2D visual features, further optimization is necessary to bridge the 2D-to-3D gap. NeRF-based methods~\cite{kerr2023-lerf, rashid2023-LERFTOGO, shen2023-F3RM} facilitate this understanding through differentiable rendering. However, NeRFs~\cite{mildenhall2021-nerf, kerr2023-lerf, rashid2023-LERFTOGO, shen2023-F3RM} are fundamentally implicit representations, making them difficult to edit to accommodate scene changes, thus leading to a {\bf static} assumption. To address dynamic problems, some works~\cite{florence2018dense,manuelli2019kpam,simeonov2022neural,yen2022nerf} commonly predict grasp poses using 3D dense correspondences, where reliable grasps are identified based on keypoints in a reference state and then applied to various viewpoints or object placements. However, these methods face challenges in tracking object states over time and handling identical objects.

To this end, we propose \acronym{}. Given posed RGBD frames from a calibrated camera, \acronym{} constructs a high-fidelity representation as a collection of explicit Gaussian ellipsoids via 3D Gaussian Splatting (3DGS)~\cite{kerbl2023-gaussiansplatting,chen2024-gssurvey}. \acronym{} reconstructs a scene in under 30 seconds and supports efficient part-level grasping for static and rigid transformation, which enables manipulation such as tracking part objects that is not possible with existing methods. \acronym{} initializes Gaussians from the coarse geometry of depth frames; while computing reference features for each input view in real time using MobileSAM~\cite{zhang2023-mobilesamv2} and MaskCLIP~\cite{zhou2022-MASKCLIP}. These Gaussians are further optimized for geometry, texture, and semantics via differentiable rasterization. The user can supply an object name query ({\it e.g., `mug'}) and part query ({\it e.g., `handle'}) for \acronym{} to efficiently predict part-level affordance and generate grasp proposals. \acronym{} directly generates grasping proposals using explicit Gaussian primitives in milliseconds, for which we extend an existing antipodal grasp generator~\cite{ten2017-gpg,gualtieri2016-gpd}. In addition, we further exploit the explicit representation to maintain high-quality representations under object displacement. Using a point tracker~\cite{karaev2023-cotracker}, \acronym{} coarsely edits the scene to capture rigid transformations and further optimizes it with {\it partial} scene reconstruction.

We implemented \acronym{} on a desktop-grade computer with a real Franka Research (FR3) robot to evaluate its efficacy in tabletop manipulation. Every component in \acronym{} is efficient and empirically runs a magnitude (\textit{$10\times$}) faster than existing work~\cite{rashid2023-LERFTOGO,shen2023-F3RM} --- computing 2D reference features, optimizing the 3D representation, and generating 2-finger grasp proposals. This makes it possible to simultaneously generate \acronym{} representation in parallel to arm scans. In experiments, \acronym{} outperforms NeRF-based methods like F3RM and LERF-TOGO, and other point-based methods.

Our contribution is threefold:

\begin{itemize}[leftmargin=3mm]
    \item {\bf A framework that advocates 3DGS for grasping representation}. \acronym{} {\bf efficiently} reconstructs scenes with geometry, texture, and semantics supervision, which outperforms baselines on zero-shot part-based grasping in terms of both accuracy and efficiency.
    \item {\bf Techniques towards an editable high-fidelity representation}, which goes beyond zero-shot manipulation in static scenes into dynamic and articulated object manipulation.
    \item {\bf Extensive real-robot experiments} that validate \acronym{} as an effective tool for zero-shot grasping in {\it both static and dynamic scenes}, which demonstrates the superiority of our method over NeRF-based or point-based methods.
\end{itemize}

\section{Related Work}
\label{sec:related}

\textbf{Language-guided Manipulation.} To support zero-shot manipulation, robots must leverage priors learned from internet-scale data. There have been some recent works~\cite{liu2024-okrobot,chen2023-nlmap-saycan,huang2023-vlmap,jatavallabhula2023-conceptfusion,yokoyama2023-vlfm,gu2023-conceptgraphs,huang2023-voxposer} that use 2D foundation vision models (CLIP~\cite{radford2021-clip}, SAM~\cite{kirillov2023-SAM}, or GroundingDINO~\cite{liu2023-groundingdino}) to build open-vocabulary 3D representations. However, these methods mostly rely on simple 2D back-projection. Without further rendering-based optimization, they generally fail to provide precise part-level information. Recently, building on the work of DFF~\cite{kobayashi2022-DFF} and LERF~\cite{kerr2023-lerf}, researchers~\cite{shen2023-F3RM,rashid2023-LERFTOGO,qiu2024-geff,wang2023-d3fields} have found that combining feature distillation with neural rendering yields promising representations for robotics manipulation, as it offers both high-quality semantics and geometry. Notably, LERF-TOGO~\cite{rashid2023-LERFTOGO} proposed conditional CLIP queries and DINO regularization for zero-shot manipulation by parts. F3RM~\cite{shen2023-F3RM} learned grasping from few-shot demonstrations. Evo-NeRF~\cite{pmlr-v205-kerr23a} focuses on NeRF specialized for stacked transparent objects, which conceptually is orthogonal to our method. However, these methods are based on NeRFs~\cite{mildenhall2021-nerf}, which is fundamentally implicit. Though certain NeRF representations can be adapted to model dynamic movement, such as grid-based methods~\cite{shen2023-F3RM}, dynamic scenes are more natural to be modeled with explicit methods.

\textbf{Grasp Pose Detection.} Grasp pose detection has been a long-standing topic~\cite{redmon2015-graspcnn,mahler2017-dexnet20,fang2020-graspnet,fang2023-anygrasp,ten2017-gpg,gualtieri2016-gpd,sundermeyer2021-contactgrasp,mousavian2019-foxgraspnet,liu2024-vbc} in robotics manipulation. Existing methods can be roughly divided into two categories: end-to-end and sampling-based approaches. End-to-end methods~\cite{fang2020-graspnet,fang2023-anygrasp,mousavian2019-foxgraspnet,liu2024-vbc} offer streamlined pipelines for grasp poses and incorporating learned semantic priors ({\it e.g.,} mugs grasped by the handle). However, these methods often require the testing data modalities ({\it e.g.,} viewpoint, object category, and transformations) to match training distribution exactly. For instance, LERF-TOGO~\cite{rashid2023-LERFTOGO} resolves viewpoint variation of GraspNet~\cite{fang2020-graspnet} by generating hundreds of point clouds for input using different transformations, requiring significant computational time. Sampling-based methods~\cite{ten2017-gpg,gualtieri2016-gpd}, on the other hand, do not learn semantic priors but offer reliable and rapid results when explicit representations are available. In this work, we find that the {\it explicit Gaussian primitives} are natural to be connected to sampling-based methods~\cite{ten2017-gpg,gualtieri2016-gpd}, and features embedded in \acronym{} complements the semantic priors via language guidance. This intuitive combination allows efficient and accurate sampling of grasping poses in dynamic and cluttered environments.

\textbf{Concurrent Work.} Concurrently, multiple approaches~\cite{zhou2023-feature3dgs,qiu2024-featuresplatting,qin2023-langsplat,zheng2024-gaussiangrasper} are beginning to interface Gaussian Splatting~\cite{kerbl2023-gaussiansplatting} with 2D features. The majority of these works focus solely on appearance editing~\cite{zhou2023-feature3dgs,qiu2024-featuresplatting,qin2023-langsplat}. We built \acronym{} based on Feature Splatting~\cite{qiu2024-featuresplatting} for its engineeringly optimized implementation and further reduces the overall reconstruction time to 1/10. During the preparation of the manuscript, a concurrent work appeared~\cite{zheng2024-gaussiangrasper}. Similar to our work, \citet{zheng2024-gaussiangrasper} also combines Gaussian Splatting~\cite{kerbl2023-gaussiansplatting} with the feature distillation for grasping. However, \citet{zheng2024-gaussiangrasper} does not handle part-level queries for task-oriented manipulation~\cite{rashid2023-LERFTOGO} and still mostly focuses on static scenarios. Though they provide a brief demonstration of the potential of Gaussian primitives in handling moving objects, they still make a strong assumption --- object representations are displaced only when they are moved by the robot arm. Such an assumption falls short in more general scenarios that involve external forces ({\it e.g.,} displacement by other machines or humans). In addition, they still require costly reference feature generation. The most concurrent work~\cite{li2024object} uses Gaussian Splatting for robotic manipulation, but it only fuses data from a few fixed cameras and thus does not address part-level manipulation. GraspSplats extends Gaussian Splatting~\cite{kerbl2023-gaussiansplatting} as a promising alternative to address these issues.

\section{Efficient Manipulation with 3D Feature Splatting}
\label{sec:method}

\paragraph{Problem Formulation.} We assume a robot with a parallel gripper, a calibrated in-wrist RGBD camera, and a calibrated third-person view camera. Given a scene containing a set of objects, the objective is for the robot to grasp and lift an object via language queries ({\it e.g., `kitchen knife'}). Optionally, a part query may be further supplied to specify the part to grasp ({\it e.g., `handle'}) for task-oriented manipulation~\cite{rashid2023-LERFTOGO}. It is worth noting that, unlike previous works~\cite{shen2023-F3RM,rashid2023-LERFTOGO}, we {\it do not} assume that the scene is static. Instead, we aim to design a more generalized algorithm where part-level grasping affordance and sampling can be done continuously even with object movement.

\paragraph{Background.} The original Gaussian Splatting~\cite{kerbl2023-gaussiansplatting} (GS) focuses on novel view synthesis and is restricted to using only texture information as supervision. Several recent works~\cite{qiu2024-featuresplatting,qin2023-langsplat,zhou2023-feature3dgs} attempt to extend GS to reconstruct dense 2D features. More concretely, \acronym{} renders the depth $\hat{\mathbf{D}}$, color $\hat{\mathbf{C}}$, and the dense visual features $\hat{\mathbf{F}}$ with the splatting algorithm:
\begin{equation}
    \{\hat{\mathbf{D}}, \hat{\mathbf{F}}, \hat{\mathbf{C}}\} = \sum_{i \in N} \{\mathbf{d}_i, \mathbf{f}_i, \mathbf{c}_i\}\cdot \alpha_i\, \prod_{j = 1}^{i - 1} (1 - \alpha_j)\,,
\label{eq:feature_splatting_rasterization}
\end{equation}
where $\mathbf{d}_i$, $\mathbf{f}_i$, and $\mathbf{c}_i$ is the per-gaussian distance to the camera origin, latent feature vector, and color, $\alpha_i$ is per-gaussian opacity, and the indices $i\in N$ are in the ascending order sorted by $\mathbf{d}_i$. Following the convention~\cite{qiu2024-featuresplatting}, we further assume that per-gaussian feature vector $\mathbf{f}_i$ is isotropic. The rendered depth, images, and features are then supervised using L2 loss. Note that all recent works~\cite{qiu2024-featuresplatting,zhou2023-feature3dgs,qin2023-langsplat} follow a similar paradigm as Eq.~\ref{eq:feature_splatting_rasterization}. We provide complete details in the supplementary materials.

\begin{figure}
    \centering
    \includegraphics[width=1.\linewidth]{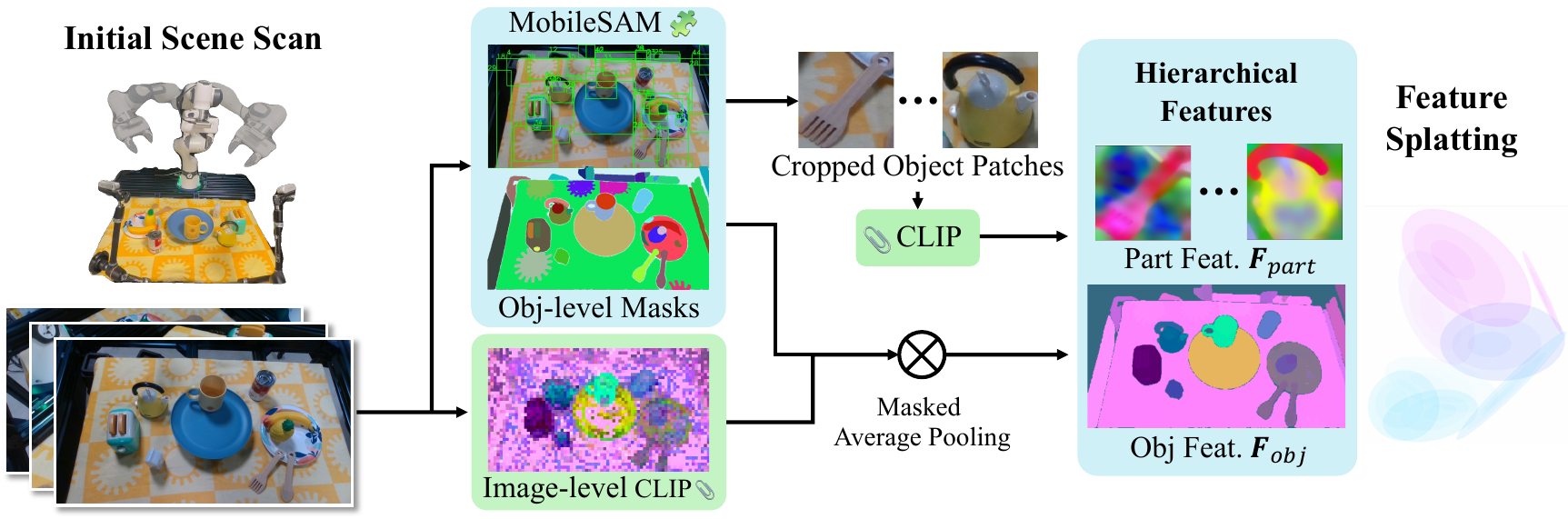}
    \caption{\acronym{} employs two techniques to efficiently construct feature-enhanced 3D Gaussians: hierarchical feature extraction and dense initialization from geometry regularization, which reduces the overall runtime to 1/10 of existing GS methods~\cite{qiu2024-featuresplatting}. (High-dimensional features are visualized using PCA and the visualized Gaussian ellipsoids are trained without densification).}
    \label{fig:feature_overview}
    \vspace{-5pt}
\end{figure}

\paragraph{Overview.} To support open-ended grasping, \acronym{} proposes three key components. The overviews are given in Fig.~\ref{fig:feature_overview} and Fig.~\ref{fig:grasp_overview}. First, a way to construct a scene representation efficiently with novel reference features and geometric regularization. Second, a way to generate grasp proposals directly on 3D Gaussians, using 3D conditional language queries and an extended antipodal grasp proposer~\cite{ten2017-gpg,gualtieri2016-gpd}. Finally, a way to edit Gaussians under object displacement which enables dynamic and articulated object manipulation.

\subsection{Constructing Feature-enhanced 3D Gaussians}

We use differentiable rasterization~\cite{kerbl2023-gaussiansplatting,qiu2024-featuresplatting} to lift 2D features to 3D representation. Though existing works in feature-enhanced GS offers part-level understanding~\cite{qiu2024-featuresplatting,qin2023-langsplat}, one commonly overlooked weakness is the expensive overhead {\it before the scene optimization begins}. This overhead can be further dissected to (1) costly reference feature computation~\cite{rashid2023-LERFTOGO} or (2) densification of sparse Gaussians~\cite{qiu2024-featuresplatting} originated from SfM preprocessing~\cite{kerbl2023-gaussiansplatting,fu2023-colmapfree}, which we address in this work.

\paragraph{Efficient Hierarchical Reference Feature Computation.} Existing methods~\cite{rashid2023-LERFTOGO,qin2023-langsplat,zheng2024-gaussiangrasper} spend most compute efforts to regularize coarse CLIP features --- either through thousands of multi-scale queries~\cite{rashid2023-LERFTOGO} or mask-based regularization~\cite{qiu2024-featuresplatting,qin2023-langsplat,zheng2024-gaussiangrasper} through costly grid sampling~\cite{kirillov2023-SAM}.

We propose a way to efficiently regularize CLIP using MobileSAMV2~\cite{zhang2023-mobilesamv2}. We generate hierarchical features, object-level and part-level, specialized for grasping. Given an input image, MobileSAMV2~\cite{zhang2023-mobilesamv2} predicts class-agnostic bounding boxes $\mathbf{D}_{obj} := \{(x_i, y_i, w_i, h_i)\}_{i = 1}^{N}$ and a set of object masks $\{\mathbf{M}\}$. For object-level feature, we first use MaskCLIP~\cite{zhou2022-MASKCLIP} to compute coarse CLIP features of the entire image $\mathbf{F}_{C} \in \mathbb{R}^{H' \times W' \times C}$. We then follow \citet{qiu2024-featuresplatting} and use Masked Average Pooling to regularize object-level CLIP features with $\{\mathbf{M}\}$, which we detail in Sec.~\ref{sec:appendix_reference_feature}.

For part-level features, we extract image patches from $\mathbf{D}_{obj}$ for batched inference on MaskCLIP~\cite{zhou2022-MASKCLIP}. Since $\mathbf{D}_{obj}$ incorporates object priors learned from the SA-1B dataset~\cite{kirillov2023-SAM}, $N$ is significantly smaller than the number of patches needed from uniform queries~\cite{kerr2023-lerf} for efficient inference. We then interpolate the features to remap them into the original image shape and average over multiple instances to form $\mathbf{F}_{part}$ for part-level supervision.

During differentiable rasterization, we introduce a shallow MLP with two output branches that takes in the rendered features $\hat{\mathbf{F}}$ from Eq.~\ref{eq:feature_splatting_rasterization} as intermediate features. The first branch renders the object-level feature $\hat{\mathbf{F}}_D$ and the second branch renders the part-level feature $\hat{\mathbf{F}}_{obj}, \hat{\mathbf{F}}_{part} = \textsc{MLP}(\hat{\mathbf{F}})$, where $\hat{\mathbf{F}}_{obj}$ and $\hat{\mathbf{F}}_{part}$ are supervised using $\mathbf{F}_{obj}$ and $\mathbf{F}_{part}$ with cosine loss. We scale the part-level term in the joint loss \( \mathcal{L}_\text{obj} + \lambda \cdot \mathcal{L}_\text{part}\,\) with $\lambda = 2.0$ to emphasize part-level segmentation.

\paragraph{Geometry Regularization via Depth.} Existing feature-enhanced GS methods~\cite{qiu2024-featuresplatting,qin2023-langsplat,zhou2023-feature3dgs} have no supervision for geometry. In \acronym{}, we project points from depth images as centers of the initial Gaussians. In addition, we use depth as supervision during training. Empirically, this additional geometric regularization significantly reduces the training time and better surface geometry.

\subsection{Static Scene: Part-level Object Localization and Grasp Sampling}

To support efficient zero-shot part-level grasping, \acronym{} performs object-level query, conditional part-level query, and grasp sampling. Unlike NeRF-based approaches~\cite{rashid2023-LERFTOGO}, which requires costly rendering to extract language-aligned features and geometry from implicit MLPs, \acronym{} operates directly on Gaussian primitives for efficient localization and grasping queries.

\textbf{Open-vocabulary Object Querying.} We first perform object-level open-vocabulary query ({\it e.g., mug}), where we take language queries to select objects for grasping, with optional negative queries to filter out other objects. We do so by directly identifying 3D Gaussians whose isotropic CLIP features more closely align with positive queries over negative queries. The feature-text comparison process follows standard CLIP practices~\cite{radford2021-clip,shen2023-F3RM} and is detailed in Sec.~\ref{ap:object_level_clip_query}.

\begin{figure}
    \centering
    \includegraphics[width=1.\linewidth]{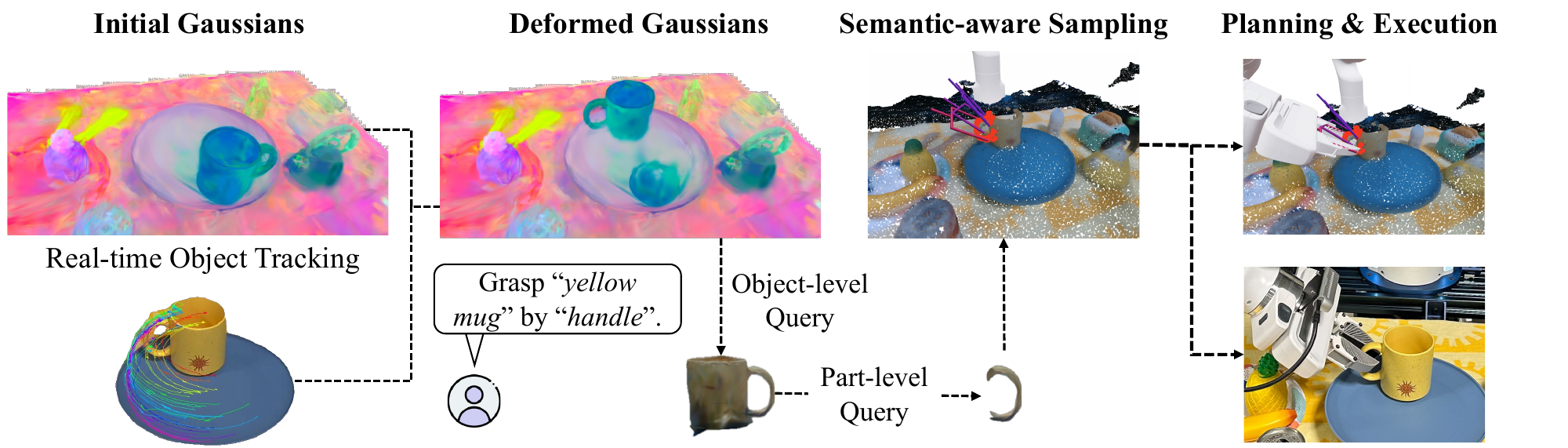}
    \caption{Given an initial state of Gaussians and RGB-D observations from one or more cameras, \acronym{} tracks the 3D motion of objects specified via language, which is used to {\it deform} the Gaussian representations in real-time. Given object-part text pairs, \acronym{} proposes grasping poses using both semantics and geometry of Gaussian primitives in milliseconds.}
    \vspace{-7pt}
    \label{fig:grasp_overview}
\end{figure}

\textbf{Open-vocabulary Conditional Part-level Querying.} As discussed by \citet{rashid2023-LERFTOGO}, CLIP exhibits bag-of-words-like behavior ({\it e.g.,} the activation of {\it `mug handle'} tends to contain both mugs and handles). Thus, it is necessary to perform {\it conditional queries}. While LERF-TOGO~\cite{rashid2023-LERFTOGO} requires a two-step (render-voxelization) process; \acronym{} {\it natively supports} CLIP queries conditioned on Gaussian primitives. In particular, given an object segmented from the previous operation, we simply repeat the procedure with the new part-level query and limit the set of Gaussians to the segmented object. A qualitative example of this part-level conditioning is given in Fig.~\ref{fig:grasp_overview}.

\textbf{Grasp Sampling using Gaussian Primitives.} We perform grasp sampling directly on the Gaussian primitives for streamlined grasping. To do so, we combine \acronym{} with GPG, a sampling-based grasp proposer~\cite{ten2017-gpg,gualtieri2016-gpd}. We first define a workspace \(\mathcal{R}_{obj}\) as the 3D space expanded from the segmented object part. The expansion radius is the sum of both the longest axes of the scales of the Gaussian primitives and the gripper's collision radius. Then, we sample $N$ points from $\mathcal{R}_{obj}$. Within the neighborhood $R_{p}$ of these sampled points where $R_{p}$ refers to the area within a specified distance from the selected point, we aggregate Gaussian primitives with rendered normals and compute the reference coordinate system for grasp sampling with the average normal direction
\begin{equation}
M(p) = \sum_{g \in R_{p}} \hat{n}(g) \hat{n}(g)^T
\label{eq:sampling_normal}
\end{equation}%
where \(\hat{n}(g)\) denotes the unit surface normal of the gaussian primitive \( g \). In the reference frame of each sampled point \(p\), we perform a local grid search to find candidate grasps, where the finger of the gripper at terminal candidate grasps contacts the geometry of the segmented part. The details are given in Sec.~\ref{ap:grasp_sampling_algo}.

\subsection{Dynamic Scene: Real-time Tracking and Optimization}
Using representations optimized for semantics and geometry, it is natural to extend \acronym{} to track object displacements and edit the Gaussian primitives in real time. It is worth noting that such operation is challenging for existing NeRF-based methods~\cite{rashid2023-LERFTOGO,shen2023-F3RM}.

\textbf{Multi-view Object Tracking with keypoints.} Assume one or more calibrated cameras without egocentric motions. Given an object language query, we segment its 3D Gaussian primitives and rasterize a 2D mask to the camera. We then discretize the rendered mask into a set of points as input to a point tracker~\cite{karaev2023-cotracker}, which continuously tracks the 2D coordinates of given points. We translate these 2D correspondences into 3D using depth. To filter out noisy correspondences, we use a simple DBSCAN~\cite{ester1996-DBSCAN} clustering algorithm to filter out 3D outliers. Finally, for the remaining correspondence points, we use the Kabsch~\cite{kabsch1976-kabsch} algorithm to solve for the $\mathbf{SE}(3)$ transformation, which we apply to the segmented 3D Gaussians primitives. For multiple cameras, we append the estimated 3D correspondences from all cameras to the system of equations for the Kabsch algorithm. Note that the displacement can be exerted either by the arm or other external forces. 

\textbf{Partial Fine-Tuning.} Edited scenes may have undesirable artifacts for regions unobserved during the initial reconstruction ({\it e.g.,} surface underneath displaced objects). Optionally, \acronym{} supports partial scene re-training using object masks rendered before and after the displacement, which is much more efficient than a complete reconstruction.


\section{Experiments}
\label{sec:result}

In this section, we conduct experiments to validate the efficacy of \acronym{}. Specifically, our experiments aim to address the following research questions:

\begin{itemize}[leftmargin=3mm]
    \item Main Results --- Why is \acronym{} preferable than existing NeRF- and point-based methods?
    \item Ablation Study --- What were the design choices?
\end{itemize}

{\bf Experiment Protocol.} We obtain calibration of all third-person cameras using Colmap~\cite{schonberger2016-colmap} with Aruco visual markers. For all settings, we first obtain an initial scan of the scene by programming the arm to go in a Bezier curve defined by given waypoints. We provide object-part text queries for all grasping experiments on the real robot. Though GraspSplats conceptually supports multi-camera tracking, we only use a single RGB-D camera for tracking in experiments. Manipulation success is defined as lifting the object (optionally, by the parts specified) for at least 3 seconds without re-attempting. We provide the protocols of static and dynamic part-level manipulation below:
\begin{itemize}[leftmargin=3mm]
    \item {\bf Static zero-shot part-level manipulation.} We experiment with 8 different re-arrangements of objects (4 are cluttered, shown in Fig.~\ref{fig:teaser_fig}) with 24 objects. We conducted 43 trials in total in this setting. The objects included kitchenware, tableware, toys, tools, and other commonly encountered items. We intentionally include many difficult cases such as part-level grasping and challenging items.
    \item {\bf Dynamic zero-shot part-level manipulation.} In the experiment, we used objects similar to static grasping (40 trials of 24 objects per type). However, a human operator rearranges the specified object after the initial scan. To specifically evaluate tracking performance, we experiment with three types of dynamic motion:
    \begin{itemize}[leftmargin=4mm]
        \item \textbf{Easy.} Objects are translated without rotation.
        \item \textbf{Medium.} Objects are rotated 180$^{\circ}$, with some occlusions from hand during rotation.
        \item \textbf{Hard.} Objects are translated and rotated simultaneously, with certain occlusions.
    \end{itemize}
\end{itemize}

\begin{table*}[t]
\centering
\begin{tabular}{l|cc|cc}
\hline
    & \multicolumn{2}{c|}{\underline{Latency$\downarrow$}} & \multicolumn{2}{c}{\underline{Grasping Success$\uparrow$}} \\ 
Method  & Training  & Grasping & Static  & Dynamic \\ \hline
Tracking Anything~\cite{cheng2023-trackAnything} & ---                   & 3.1s             & 41.9\%                                  & 45\% \\
ConceptGraphs\textcolor{red}{$^{\star}$}~\cite{gu2023-conceptgraphs} & $\sim$30s             & 0.7s             & 51.1\%                                  & ---\textcolor{red}{$^{\dagger}$} \\
LERF-TOGO~\cite{rashid2023-LERFTOGO}             & $\sim$10min           & 9.9s             & 65.1\%                                  & ---\textcolor{red}{$^{\dagger}$} \\
F3RM\textcolor{red}{$^{\star}$}~\cite{shen2023-F3RM}                  & $\sim$3min            & 1.6s             & 72.1\%                                  & ---\textcolor{red}{$^{\dagger}$}   \\ 
\acronym{} (Ours)                                     & \textbf{60s}          & 1.3s             & \textbf{81.4\%}                         & \textbf{74.2\%} \\
\hline
\end{tabular} 
\caption{\textbf{Comparison to NeRF and 2D-based methods on latency, static/dynamic successful rate.} Latencies are reported for (1) the time for (re)building the representation; and (2) grasp latency given task texts. The reported success rates are averages of variations in motions and objects/parts. \textcolor{red}{$^{\star}$}: reproduced variants that use GraspNet~\cite{fang2020-graspnet,fang2023-anygrasp} on objects segmented by the underlying methods. \textcolor{red}{$^{\dagger}$}: methods require offline batch processing that does not cope with dynamic scenes.}
\vspace{-5pt}
\label{table:main_results}
\end{table*}

\subsection{Main Results}
In the main results section, we compare GraspSplats with strong baselines including NeRF-based~\cite{rashid2023-LERFTOGO,shen2023-F3RM} methods, 2D+Depth point-cloud based method~\cite{cheng2023-trackAnything}, and scene-graph based method~\cite{gu2023-conceptgraphs}. Specifically, We compare with two recent NeRF-based methods, LERF-TOGO~\cite{rashid2023-LERFTOGO} and F3RM~\cite{shen2023-F3RM}. Since F3RM requires human demonstrations, we implement a zero-shot variant, F3RM$^{*}$. F3RM$^{*}$ renders CLIP features and depth for segmentation and geometry, which uses GraspNet~\cite{fang2020-graspnet} to generate grasps. In addition, for the 2D baseline, we use TrackAnything~\cite{cheng2023-trackAnything}, which combines SAM~\cite{kirillov2023-SAM} and GroundingDINO~\cite{liu2023-groundingdino} for segmentation. The foreground depth and images are used for building point-cloud, then jointly fused as input for GraspNet~\cite{fang2020-graspnet} to generate grasps. For the scene-graph-based method, we use ConceptGraphs~\cite{gu2023-conceptgraphs} that leverages vision foundation models to build interactable scene graphs for manipulation. The results for latency, and success rate for static and dynamic scenes are shown in Table.~\ref{table:main_results}. We claim the following advantages for \acronym{}:

\textbf{Our approach is extremely faster than the NeRF-based methods.} As shown in Table.~\ref{table:main_results}, both our training latency (the time for rebuilding the representation) and grasping latency are faster than LERF-TOGO and F3RM. Specifically, our training latency is 10x faster than LERF-TOGO, and 3x faster than F3RM. While with supreme efficiency, our approach also has a better static scene success rate with 16.3 points better than LERF-TOGO, and 9.3 points better than F3RM.  

\textbf{With comparable latency with 2D and Scene-Graph based method, our approach doubled the success rate}. 2D-based approach is usually fast and generalizable while using 2D perception and depth map to build the point cloud for grasping and manipulation. Here, we compared with baseline TrackAnything powered by GraspNet for manipulation in both static and dynamic scenes. In static scenes, TrackAnything fails on almost every part-level grasping trials, as its internal detection model~\cite{liu2023-groundingdino} was not trained on part-level data. In dynamic scenes, its success rate is approximately same across different motion setups, as it is based on 2D object-level segmentation which again fails on part-level experiments. The scene-graph-based method generates structured object-level representations. By fusing multi-view images, it achieves slightly higher performance than TrackAnything~\cite{cheng2023-trackAnything}, but still 21 points lower than our approach.

\textbf{Our method has supreme capabilities for accurate dynamic scene modeling.} GraspSplats exploits the benefit of explicit representations, which allows editing reconstructed representations without compromising the rich geometric and semantic scene features. We demonstrate that our real-time tracking algorithm allows displacing reconstructed objects effectively, which works in dynamic scenes with a high success rate. This would otherwise be a challenging scenario for implicit methods. Additionally, our approach reconstructs scenes with greater detail than 2D-based methods, resulting in a higher success rate in dynamic scenarios.

\begin{table}[t]
  \centering
  \begin{minipage}[t]{0.51\textwidth}
    \centering
    \resizebox{1.0\textwidth}{!}{%
      \begin{tabular}{c c c}
      \toprule
      Method & Segment & Grasp Sampling \\
      \midrule
      Tracking Anything\textcolor{red}{$^{\star}$} & 2.5 $\pm$ 0.1s & $0.6 \pm 0.05$s \\
      ConceptGraph\textcolor{red}{$^{\star}$} & \textbf{0.1} $\pm$ 0.05s & $0.6 \pm 0.05$s \\
      LERF-TOGO & $5.1 \pm 0.3$s & $4.8 \pm 0.7$s \\
      F3RM & $1.0 \pm 0.1$s & $6.9 \pm 0.45$s \\
      GraspSplats & $0.8 \pm 0.1$s & \textbf{0.5} $\pm$ 0.06s \\
      \bottomrule
      \end{tabular}%
    }
    \caption{\textbf{Grasping latency} breakdown. standard deviations are reported over 10 runs. \textcolor{red}{$^{\star}$}: our own reproductions that use GPG~\cite{ten2017-gpg} for grasping.}
    \label{table:grasp_sampling_time_breakdown}
  \end{minipage}
  \hfill
  \begin{minipage}[t]{0.45\textwidth}
  \vspace{-0.5in}
    \centering
    \resizebox{1.\textwidth}{!}{%
      \begin{tabular}{c c c}
      \toprule
      Method & Query Time$\downarrow$ & Succ.$\uparrow$ \\
      \midrule
      GraspNet-100~\cite{rashid2023-LERFTOGO} & 10.3s & 76.7\% \\
      GraspNet-1~\cite{fang2020-graspnet} & 0.6s & 65.1\% \\
      F3RM~\cite{shen2023-F3RM} & 6.9s & --- \\
      \textbf{GraspSplats} & \textbf{0.5s} & \textbf{81.4\%} \\
      \bottomrule
      \end{tabular}%
    }
    \caption{\textbf{Query time and success rate} of different grasp sampling methods measured in the static scene. LERF-TOGO~\cite{rashid2023-LERFTOGO} uses multi-view inferences that require 100 GraspNet passes.
    }
    \label{table:grasp_sampling}
  \end{minipage}
\vspace{-23pt}
\end{table}
\subsection{Empirical Insights}

\textbf{Comparison to 2D segmentation.} While we empirically observe that 2D open-vocabulary segmentation models~\cite{liu2023-groundingdino,ren2024-groundedSAM,cheng2023-trackAnything} achieve similar object-level segmentation accuracy to \acronym{}, they fall short in {\bf part-level segmentation}, which is crucial for part-level task-oriented manipulation ({\it e.g.,} TrackAnything~\cite{cheng2023-trackAnything} segments mugs but often fail to segment {\it handle} of the mug). In addition, without a consistent 3D representation, it fails if the part to manipulate is out of sight to the camera.

\textbf{Comparison to NeRF-based methods.} Although NeRF-based methods demonstrate better performance on static scenes than 2D projection methods, they are \textbf{fundamentally implicit}, which leads to failure when objects are displaced by external forces. In addition, the implicit representation also requires costly training from random initialization, slow segmentation, and grasping query time. Finally, F3RM~\cite{shen2023-F3RM} and LERF-TOGO~\cite{rashid2023-LERFTOGO} fall short under cluttered scenes with object overlapping without regularizing the object boundary in the reference feature.

\subsection{Ablation Study}
\begin{floatingtable}[r]{
  \begin{tabular}{c|c}
  \hline
  \small{Method} & \small{IoU$\uparrow$} \\
  \hline
  \small{LERF~\cite{kerr2023-lerf}} & \small{39.0} \\
  \small{\textbf{GraspSplats}} & \small{\textbf{50.7}} \\
  \hline
  \end{tabular}
}
\vspace{-5pt}
\caption{\small{object/part IoU}}
\label{table:segmentation_acc}
\end{floatingtable}

\textbf{Reference Feature.} As shown in Table.~\ref{table:segmentation_acc}, we verify the effectiveness of our hierarchical features by using the mean IoU of 2D open-vocabulary segmentation. Specifically, since there are no existing scene-scale part-level segmentation datasets, we manually annotate object-level and part-level masks for 4 scenes and render CLIP features for binary masks with MaskCLIP~\cite{zhou2022-MASKCLIP}. The results in Table.~\ref{table:segmentation_acc} show that \acronym{} outperforms LERF~\cite{kerr2023-lerf} on both object-level and part-level segmentation on average with 11.7 points higher on object IoU.

\textbf{Grasp Sampling.} We ablate different grasping methods in Table.~\ref{table:grasp_sampling}. GraspNet-n represents the grasps accumulated after running GraspNet~\cite{fang2020-graspnet} n times on point clouds from different viewpoints, following the approach in the work\cite{rashid2023-LERFTOGO}. GraspSplats demonstrates superior stability and speed compared to GraspNet, owing to its reliance on sampling methods. The research finding is that, if the semantic affordance can be satisfactorily provided by the representation, then only geometrically informed sampling needs to be performed. In addition, unlike end-to-end methodologies~\cite{fang2020-graspnet}, GraspSplats consistently retains crucial grasps even in complex scenes with incomplete 3D geometry. We report the grasp optimization time F3RM~\cite{shen2023-F3RM} using its reported number; hence, since demonstrations are required, the success rate of F3RM~\cite{shen2023-F3RM} is not applicable. 

Meanwhile, we show the grasp sampling breakdown for segmentation and grasping in Table.~\ref{table:grasp_sampling_time_breakdown}. Segmentation time represents the time from object/part text queries are provided to select the correct part for grasping and grasp sampling time represents grasp pose selection after the part is localized. It is clearly shown in the Table.~\ref{table:grasp_sampling_time_breakdown}, our approach is very efficient in both segmentation and grasping.

\textbf{Initialization from Depth.} In Table.~\ref{table:initialization_scheme}, we ablate the effect of geometric regularization by comparing several variants to initialize geometry for Gaussian. In particular, while the original Gaussian Splatting~\cite{kerbl2023-gaussiansplatting} paper adopts sparse Colmap~\cite{schonberger2016-colmap} initialization from RGB images (Colmap-S); some recent works have found that dense Colmap reconstruction (Colmap-D) serves as a better initialization that results in faster convergence~\cite{wu2023-4dgs}. \acronym{} adopt geometric regularization and directly initializes centers of Gaussians using depth inputs, which leads to much more efficient training.

\textbf{Success Rate Breakdown.} As shown in Table.~\ref{table:object_part_breakdown}, to more clearly demonstrate the comparison of part-level segmentation capabilities, we tested the success rates of 27 object-part pairs in unclustered scenes, which differs from Table. ~\ref{table:main_results}. It clearly shows that our approach has nearly a 100\% success rate on object-level grasping while maintaining a very high success rate on part-level grasping. This is a benefit from both the feature-enhanced 3D Gaussians and part-level supervision from SAM.

\begin{table}[t]
  \centering
  \begin{minipage}[t]{0.5\textwidth}
    \centering
    \resizebox{\textwidth}{!}{%
      \begin{tabular}{c c c}
      \toprule
      Method & Process Time$\downarrow$ & Train Iteration $\downarrow$ \\
      \midrule
      Colmap-S~\cite{kerbl2023-gaussiansplatting} & 11.6 & 10,000 \\
      Colmap-D~\cite{wu2023-4dgs} & 623.0 & 3,000 \\
      \textbf{GraspSplats} & \textbf{0.7} & \textbf{3,000} \\
      \bottomrule
      \end{tabular}%
    }
    \caption{Comparison of different initialization schemes. Processing time is averaged across 4 scenes. Train iteration reports the updates needed for convergence in increments of 1,000.}
    \label{table:initialization_scheme}
  \end{minipage}
  \hfill
  \begin{minipage}[t]{0.45\textwidth}
    \vspace{-0.37in}
    \centering
    \resizebox{\textwidth}{!}{%
      \begin{tabular}{c c c}
      \toprule
      Method & Object-level & Part-level \\
      \midrule
      LERF-TOGO~\cite{rashid2023-LERFTOGO} & 81.5\% & 63.0\% \\
      F3RM\textcolor{red}{$^{\star}$}~\cite{shen2023-F3RM} & 85.2\% & 77.8\% \\
      \textbf{GraspSplats} & \textbf{96.3\%} & \textbf{85.2\%} \\
      \bottomrule
      \end{tabular}%
    }
    \caption{Analysis of object-level and part-level \textbf{grasping success rate} under static scenes. Note that this table is computed using 27 object-part pairs different from the runs presented in Table.~\ref{table:main_results}.}
    \label{table:object_part_breakdown}
  \end{minipage}
\vspace{-20pt}
\end{table}

\vspace{3pt}
\subsection{Qualitative Results}
\vspace{-2pt}

\begin{figure*}[t!]
    \centering
    \includegraphics[width=\linewidth]{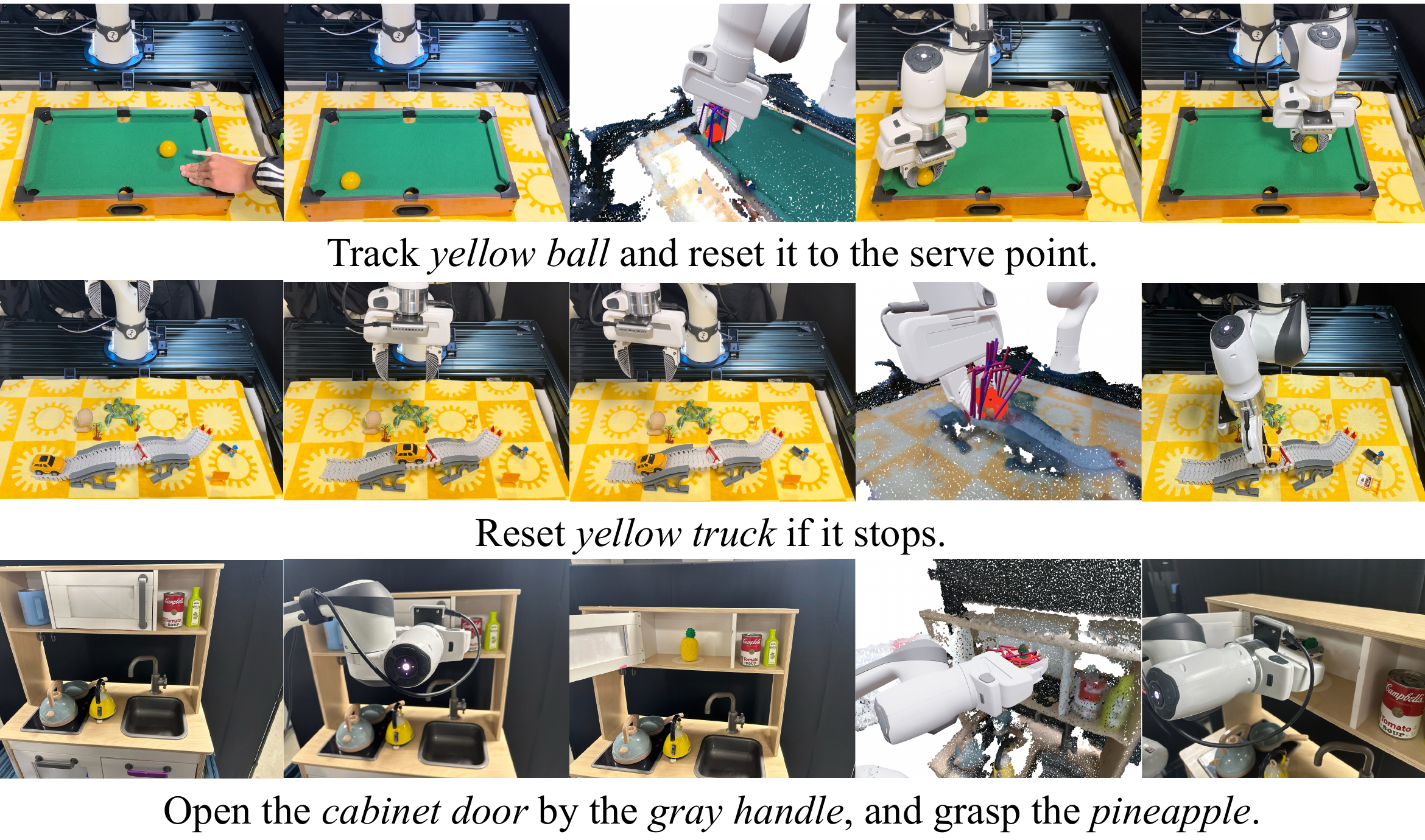}
    \vspace{-10pt}
    \caption{Qualtative examples of \acronym{} performing zero-shot task execution in real-world environments. Given object-part text queries (\textit{italicized} in the task description), \acronym{} executes grasping followed by heuristic trajectories. From left to right each row: illustration of scene change; grasp poses sampled by GraspSplats; execution of grasping. Animated visualization can be found on the website.}
    \label{fig:qualitative_results}
\vspace{-10pt}
\end{figure*}

Fig.~\ref{fig:qualitative_results} demonstrates the qualitative performance of the GraspSplats system in executing zero-shot tasks in real-world environments. The tasks involve various object-part text queries, such as tracking a yellow ball, resetting a yellow truck, and opening a cabinet door to grasp a pineapple. GraspSplats effectively samples grasp poses and executes grasping and heuristic trajectories based on these queries. The images sequentially showcase the scene change, grasp pose sampling, and the successful execution of the required task, highlighting the system’s ability to perform complex manipulations in diverse environments without prior specific training on these tasks.

\vspace{-5pt}
\subsection{Failure Cases Analysis}
\vspace{-2pt}

In static scenes, the system effectively segments objects, though challenges arise when objects are very similar, leading to segmentation failures. Most task failures stem from execution issues, like collisions during lifting, which require more advanced planning beyond our current scope. Additionally, while the system can place objects with the same orientation as grasped, real-world scenarios often involve object rotation during grasping, complicating the placement due to limited perception caused by the gripper obstructing the view.
In dynamic scenes, the success of grasping relies heavily on accurate tracking. Objects with complex textures and asymmetry are easier to track, while single-color or symmetric objects pose significant challenges, often resulting in tracking failures during long-term or rapid movements. Enhancing tracking accuracy, especially through re-sampling keypoints during tasks, will be a focus of future research.

\section{Conclusion}
\label{sec:conclusion}

In this work, we present \acronym{}, a novel representation for efficient part-based zero-shot manipulation. Using hierarchical features, geometric regularization, and grasp sampling, \acronym{} are efficient in both optimizing feature-enhanced 3D representations and supporting grasp queries via language. We further enhance \acronym{} with point trackers to directly edit optimized representation to capture the dynamics of objects --- outperform implicit NeRF-based methods. The results are validated on a real robot for tabletop manipulation under diverse settings.

{\bf Limitations.} We highlight the most pronounced limitations to facilitate future research. The current object corresponding algorithm based on the Kabsch~\cite{kabsch1976-kabsch} algorithm assumes that objects undergo {\it rigid} transformation. While the Gaussian representation is conceptually applicable to more general deformable objects ({\it e.g.,} dough and clay), this is not investigated in the current work. In addition, the tracking is sensitive to fast rotation and the resulting visual occlusions and motion blurs, which could be potentially addressed as an optimization problem using semantic and geometric priors fused in \acronym{} without assuming consistent object views.




\bibliography{main}  

\clearpage

\appendix

\section{Part-level Features Determination}
Given object-level bounding boxes, we crop and wrap images to square patches to obtain CLIP features. After per-patch features are computed, we create a part-level feature map by aggregating per-patch features using the inverse function of the cropping and wrapping process. For pixels that are assigned with more than two feature vectors, the features are averaged. Pixels with no assigned features are ignored during rendering-based feature distillation.

\section{Object-level CLIP Queries}
\label{ap:object_level_clip_query}

Specifically, \acronym{} follows standard CLIP querying practices~\cite{shen2023-F3RM,kerr2023-lerf,rashid2023-LERFTOGO,qiu2024-featuresplatting} and takes a positive vocabulary with negative vocabularies. By default, the negative vocabularies include canonical words ({\it i.e., `objects' and `things'}), but can be optionally extended with user queries. To be specific, given a language set   $L = \{L_0^-, L_1^-, \dots, L_n^-, L^+\}$    containing $n+1$ words, where $L^+$ is the positive query and the remaining $L^-$ are negative queries. The CLIP model is used to encode each word $L_i$ into a 768-dimensional feature vector $\mathbf{F}_{\text{text}, i} \in \mathbb{R}^{768}$:

$$
\mathbf{F}_{\text{text}, i} = \text{CLIP\_encode}(L_i), \quad i = 0, 1, \dots, n
$$

For each Gaussian primitive $j$, there is a 16-dimensional view-independent feature vector $\mathbf{F}_{\text{latent}, j} \in \mathbb{R}^{16}$ This is decoded into a 768-dimensional CLIP feature representation $\mathbf{F}_{\text{CLIP}, j} \in \mathbb{R}^{768}$:

$$
\mathbf{F}_{\text{CLIP}, j} = \text{Decoder}(\mathbf{F}_{\text{latent}, j})
$$

Then, we calculate the cosine similarity between its CLIP feature representation $\mathbf{F}_{\text{CLIP}, j}$ and each query $L_i$ in the set $L$ :

$$
\text{sim}(\mathbf{F}_{\text{CLIP}, j}, \mathbf{F}_{\text{text}, i}) = \frac{\mathbf{F}_{\text{CLIP}, j} \cdot \mathbf{F}_{\text{text}, i}}{\|\mathbf{F}_{\text{CLIP}, j}\| \|\mathbf{F}_{\text{text}, i}\|}
$$

Then, we apply a softmax function to the similarities between it and all queries  $L_i$ to enhance the similarity for the positive query:

$$
\mathbf{S}_{j} = \text{softmax}(\{\text{sim}(\mathbf{F}_{\text{CLIP}, j}, \mathbf{F}_{\text{text}, i})\}^{n}_{i=0})
$$

Here, $\mathbf{S}_{j}$ is the similarity vector for Gaussian primitive $j$. After applying a temperature softmax, the similarity for the positive query $L^+$ is selected:

$$
\text{sim}_{\text{positive}, j} = \mathbf{S}_{j}[n]
$$

The selecting Gaussians whose similarity to $L^+$ passes a certain threshold $\tau = 0.6$ will be regarded as the grasping object. After Gaussians are selected, we apply the DBSCAN~\cite{ester1996-DBSCAN} clustering algorithm to filter out outliers.

\section{Reference Feature Computation}
\label{sec:appendix_reference_feature}

Following \cite{zhang2023-mobilesamv2}, $\mathbf{D}_{obj}$ achieves high recall by intentionally including more false positives, which ensures recall of objects. MobileSAMV2~\cite{zhang2023-mobilesamv2} then uses $\mathbf{D}_{obj}$ as priors to generate object-level segmentation masks \(\{\mathbf{M}\}\).

To generate object-level features, For a given object mask $\mathbf{M}$, we use Masked Average Pooling (MAP) to aggregate an object-level feature vector%
\begin{equation}
    w_i = \textsc{MAP}(\mathbf{M}, \mathbf{F}_{C}) = \frac{\sum_{i \in \mathbf{F}_{C}} \mathbf{M}(i) \cdot \frac{\mathbf{F}_{C}(i)}{\left| \left| \mathbf{F}_{C}(i) \right| \right|}}{\sum_{i \in \mathbf{F}_{C}} \mathbf{M}(i)}\,,
\end{equation}%
where $i$ is a pixel coordinate. We then construct $\mathbf{F}_{obj}$ by assigning ${w}$.

To generate part-level features, for a given bounding box, we crop and wrap the patches to (224, 224). The image patch is then processed by MaskCLIP to generate feature map of shape (28, 28, 768). We then interpolate the generated feature map to match the size of the original bounding box, and paste it onto the part-level feature map. If a pixel is assigned more than one feature, we average all assigned features.

\section{Grasp Sampling Algorithm}
\label{ap:grasp_sampling_algo}

We define \( F(p) = [v_3(p) v_2(p) v_1(p)] \) as the orthogonal reference frame at point \( p \) where \( v_1(p), v_2(p), v_3(p) \), correspond to the normal direction, the secondary direction, and the minimum direction of \( M(p)\).  We search a 2D grid \( G = Y \times \Phi \). For each \( (y, \phi) \in G \), we apply translations and rotations relative to \( F(p) \), then push the gripper along the negative x-axis until a finger or the base of the gripper contacts the point cloud. Let \( T_{x,y,\phi} \) be the homogeneous transformation describing translations in the \( x,y \) plane and rotation about the z-axis. The gripper \( h_{y,\phi} \) at grid cell \( y, \phi \) contacts the point cloud at: \( F(h_{y,\phi}) = F(p)T_{x^*,y,\phi} \), where \( x^* \) is the minimum distance along the negative x-axis at which the gripper contacts the point cloud. If the number of segmented objects \( N_{obj} \) within the gripper's closed region exceeds a set threshold \( N_{th} \), i.e., \( N_{obj} > N_{th} \), the gripper \( h_{y,\phi} \) is added to the candidate grasp set \( H \). Finally, we use a geometry-aware scoring model~\cite{gualtieri2016-gpd,singh2014-bigbird} to rank the grasps and select the grasp pose with the highest score.

\section{Analysis of Failure Cases of \acronym{}.}

Though \acronym{} achieves an overall good success rate in static scenes, its performance degrades as the object becomes more dynamic. We find that \acronym{} copes with translation motion well. Empirically, the speed at which the object is translated does not significantly impact the accuracy of the deformation. \acronym{} degrades when the object undergoes drastic rotation, which causes occlusions and loss of correspondence.

\section{Hardware Configuration}

We use the Franka Research robot (FR3) as the main experiment platform. In addition to one Intel Realsense D435 cameras mounted on the end effector, we also use inputs from two third-person view Intel Realsense D435 cameras to facilitate scene reconstruction. We use the UMI gripper~\cite{chi2024-umigripper} as the end effector. The arm and cameras are connected to a desktop computer with a single RTX 4090 and an Intel i9-13900k CPU, on which \acronym{} is deployed.

\section{Multi-state Task Policy Rollout}
The examples of complex tasks are designed to illustrate potential use cases rather than showcase highly intricate operations. The tasks primarily focus on object manipulation, such as updating an object's position, grasping, resetting, and pick-and-place actions. For instance, in the toy car experiment in the supplementary video, after grasping the car, the robot resets the car by placing it back in its original position with the same orientation. During pick-and-place tasks, the placement orientation is aligned with the grasping orientation, and the placement location is determined by querying the center of another object.

These operations highlight the system's capabilities in basic object manipulation. However, challenges arise in tasks that involve precise placement, especially when the object's orientation shifts during grasping due to the inherent instability or rotation caused by the gripper's contact. This issue is particularly evident when the gripper obstructs the view of the object, complicating the perception of the object's orientation during placement. These challenges and potential failure cases are further discussed in the supplementary material.

\end{document}